**ConvPath: A Software Tool for Lung Adenocarcinoma Digital Pathological Image Analysis Aided by Convolutional Neural Network**

Shidan Wang[1,#]; Tao Wang, PhD[1,2,#]; Lin Yang, MD[1,3,#]; Faliu Yi, PhD[1]; Xin Luo, PhD[1]; Yikun Yang, MD[4]; Adi Gazdar, MD[5,8]; Junya Fujimoto, MD[6]; Ignacio I. Wistuba, MD[6]; Bo Yao, PhD[1]; ShinYi Lin, MS[1]; Yang Xie, MD, PhD[1,7,8]; Yousheng Mao, MD[4]; Guanghua Xiao, PhD[1,7,8,*]

[1]Quantitative Biomedical Research Center, Department of Clinical Sciences, University of Texas Southwestern Medical Center, Dallas, TX

[2]Center for the Genetics of Host Defense, University of Texas Southwestern Medical Center, Dallas, TX

[3]Department of Pathology, National Cancer Center/Cancer Hospital, Chinese Academy of Medical Sciences (CHCAMS), China

[4]Department of Thoracic Surgery, National Cancer Center/Cancer Hospital, Chinese Academy of Medical Sciences (CHCAMS), China

[5]Department of Pathology, University of Texas Southwestern Medical Center, Dallas, TX

[6]Department of Translational Molecular Pathology, University of Texas MD Anderson Cancer Center, Houston, TX

[7]Department of Bioinformatics, University of Texas Southwestern Medical Center, Dallas, TX

[8]Harold C. Simmons Comprehensive Cancer Center, University of Texas Southwestern Medical Center, Dallas, TX

# These authors contributed equally to this work.

**Corresponding Author:**

Guanghua Xiao, PhD, Quantitative Biomedical Research Center, Department of Clinical Sciences, Harold C. Simmons Comprehensive Cancer Center, UT Southwestern Medical Center, Dallas, TX 75390; e-mail: Guanghua.Xiao@UTSouthwestern.edu

**Running title:** ConvPath for Digital Pathological Image Analysis

**ABSTRACT**

The spatial distributions of different types of cells could reveal a cancer cell's growth pattern, its relationships with the tumor microenvironment and the immune response of the body, all of which represent key "hallmarks of cancer". However, manually recognizing and localizing all the cells in pathology slides is almost impossible. In this study, we developed an automated cell type classification pipeline, ConvPath, which includes nuclei segmentation, convolutional neural network-based tumor/stroma/lymphocytes classification, and extraction of tumor microenvironment-related features for lung cancer pathology images. The overall classification accuracy is 92.9% and 90.1% in training and independent testing datasets, respectively. By identifying cells and classifying cell types, this pipeline can convert a pathology image into a "spatial map" of tumor/stroma/lymphocyte cells. From this spatial map, we can extract features that characterize the tumor micro-environment. Based on these features, we developed an image feature-based prognostic model and validated the model in two independent cohorts. The predicted risk group serves as an independent prognostic factor, after adjusting for clinical variables that include age, gender, smoking status, and stage. ConvPath software is a user-friendly tool for pathologists and bioinformaticians and is available at https://qbrc.swmed.edu/projects/cnn/.

**Word Count:** 137

Lung cancer is the leading cause of death from cancer, in the United States as well as worldwide. Lung adenocarcinoma (ADC) accounts for almost 50% of primary lung malignancies and has remarkable heterogeneity in clinical, radiologic, molecular and pathologic features[1]. The new 2015 World Health Organization (WHO) histological classifications explicate several subtypes[2, 3]. The prognostic and predictive significance of the new ADC subtype classification has been verified by pathologist observation in surgical specimens[4, 5]. However, separating different ADC subtypes requires the pathologist to distinguish the subtle morphological patterns of pathology slides. This is time-consuming, subjective, and generates considerable inter- and intra-observer variation, even with experienced pathologists[6].

With the advance of technology, tumor tissue slide scanning is becoming a routine clinical procedure and can produce massive pathological images that capture histological details in high resolution. Tumor tissue pathology images not only contain essential information for tumor grade and subtype classifications[2], but also information on tumor microenvironment and the spatial distributions of and interactions among different types of cells. The major cell types in a malignant tissue of lung include tumor cells, stroma cells, and lymphocytes. Stromal cells are connective tissue cells such as fibroblasts and pericytes, and their interaction with tumor cells plays an important role in cancer progression[7-9] and metastasis inhibition[10]. Tumor-infiltrating lymphocytes are white blood cells that have migrated into a tumor. They are a mix of different types of cells, with T cells being the most abundant population. Tumor-infiltrating lymphocytes have been associated with patient prognosis in multiple tumor types[11-14]. Important information can be derived from cell-based image features, such as cell type, cell counts, cell spatial distributions and cell-cell interactions. One previous study using this type of feature is by Yuan et al[15], which discovered that lymphocyte percentages and spatial distribution patterns are associated with patient survival. In addition, a more recent study[16] used convolutional neural network (CNN) to identify lymphocytes and showed that the spatial organization of tumor-infiltrating lymphocytes (TILs) is associated with patient survival outcome.

The spatial distributions of different types of cells could reveal a cancer cell's growth pattern, its relationships with the tumor microenvironment and the immune response of the body, all of which represent key "hallmarks of cancer". Cell identification and classification in tumor pathology imaging could greatly facilitate the study of cell spatial distributions and their roles in tumor progression and metastasis. However, it is impractical for a pathologist to manually recognize and localize every individual cell in a pathology slide. Automated cancer pathology image recognition systems have been previously used for cancer detection[17] and prognostic feature extraction[18, 19]. Deep learning is a modern branch of machine learning; CNN, one of several deep learning strategies, performs overwhelmingly in image recognition[20-22]. CNNs have been recently applied in pathology imaging to identify lymphocyte cells[16]. However, a deep learning system which can automatically distinguish tumor, stromal, and lymphocyte cells has not been developed yet.

In this study, an automated image analysis pipeline, ConvPath, was developed for lung ADC pathology images. It includes nuclei segmentation, cell type recognition using CNN, and extraction of tumor microenvironment-related features (**Figure 1**). Pathology imaging slides and clinical information used in this study were obtained from The Cancer Genome Atlas lung ADC project (the TCGA dataset), the National Lung Screening Trial project (the NLST dataset), the University of Texas Special Program of Research Excellence (SPORE) in Lung Cancer project (the SPORE dataset), and the National Cancer Center/Cancer Hospital of Chinese Academy of Medical Sciences, China (the CHCAMS dataset). A prognostic model based on extracted cell-level image features was developed. The predicted risk score is predictive of overall survival and tumor recurrence independent of other clinical features.

## METHODS

### Datasets

H&E-stained histological images for lung ADC patients and corresponding clinical data were collected from four independent cohorts (NLST, TCGA, SPORE, and CHCAMS). The TCGA data, including 1337 tumor images from 523 patients, were obtained from the TCGA image portal (https://wiki.cancerimagingarchive.net/display/Public/TCGA-LUAD). All TCGA images were captured at X20 or X40 magnification and included both frozen and Formalin-Fixed, Paraffin-Embedded (FFPE) slides. The NLST data, including 345 tumor images from 201 patients, were acquired from the National Lung Screening Trial, which was performed by the NIH National Cancer Institute. All NLST images were FFPE slides and captured at 40X magnification. The CHCAMS data, including 102 images from 102 stage I ADC patients, were obtained from the National Cancer Center/Cancer Hospital, Chinese Academy of Medical Sciences and Peking Union Medical College (CHCAMS), China. All CHCAMS images were FFPE slides and captured at 20X magnification. The SPORE data, including 130 images from 112 patients, were acquired from the UT Lung SPORE tissue bank. All SPORE images were FFPE slides and captured at 20X magnification. The characteristics of the four datasets used in this study are summarized in **Supplemental Table 3**.

### Extraction of image patches centering at nuclei centroids

A pathologist, Dr. Lin Yang, reviewed the H&E-stained pathology image slides and manually labeled ROI boundaries using the annotation tool of ImageScope (Leica Biosystem, **Figure**

2a). ROIs were defined by the main malignant area within the pathology images. ConvPath randomly selected 10 sampling regions from each selected ROI. The sampling regions were sized 5000×5000 or 2500×2500 pixels in 40X or 20X magnification images, respectively. In each sampling region, ConvPath further extracted 80×80 image patches centering at nuclei centroids (**Figure 2b, Supplemental Figure 1**).

In order to extract the image patches, RGB color space was first converted to H&E color space with the deconvolution matrix set as [0.550 0.758 0.351; 0.398 0.634 0.600; 0.754 0.077 0.652][23]. Morphological operations consisting of opening and closing were adopted to process the hematoxylin channel image[24]. Then, ConvPath detected nuclei boundaries using a level set segmentation technique[25]. In this segmentation method, the initial contour was randomly given, the value of sigma in Gaussian filter was 1, the number of iterations was 30, and the velocity term was 60. Next, nuclei centroids were detected as the moment centroids of connected targets in a binary image, where the foreground was the regional maximum locations in a distance map of the segmented image. Here, Euclidean distance was utilized for the distance transform and regional maximums were searched within 8-connected neighborhoods. Finally, image patches using the detected nuclei centroids as centers were extracted from the original pathological RGB image (**Figure 2b, Supplemental Figure 1**).

**Deep learning algorithm in the ConvPath software**

ConvPath incorporates a CNN[26-28] to recognize the major cell types, including tumor cells, stroma cells and lymphocytes, in the center of pathology image patches (**Figure 3a**). The input to the CNN was an 80x80 image patch normalized to the range [-0.5, 0.5] with 3 channels corresponding to the red (R), green (G), and blue (B) channels. The output layer for the CNN was a softmax layer with 3 categories: tumor cell, stroma cell, and lymphocyte. For one image patch, a probability for each of the 3 categories was predicted by the CNN; the category with the highest probability was assigned as the predicted class for the image patch. The CNN was trained using a batch size of 10, a momentum of 0.9, a weight decay of 0.0001, an initial learning rate of 0.01, which shrinks by 0.99995 in each step, and training steps of 20,000. The image patches were rotated and flipped to augment sample size. A drop connect probability of 0.5 was used in all convolutional layer parameters. The NLST and TCGA datasets were combined and used as the training set for the CNN (**Figure 3 b&c**, **Supplemental Table 2**), and the SPORE dataset was used as the external validation set. The image patches in training and validation sets were labeled by the pathologist as ground truth.

**Tumor micro-environment feature extraction.**

Based on the prediction results of the CNN, ConvPath converted the pathology image into a "spatial map" of tumor cells, stromal cell and lymphocyte. From this spatial map, we can define tumor cells, stromal cell and lymphocyte regions, and characterize the distribution and interactions among these regions. Specifically, ConvPath used kernel smoothers to define regions of tumor cells, stromal cell and lymphocyte separately within the ROI (**Figure 4b**). For instance, to define the tumor cell region, ConvPath extracted coordinates of the center of all image patches and labeled them as 1 if they had been recognized as tumor cells from the previous step, 0 if not. For each point on the image, ConvPath then calculated the probability of being a tumor cell region by weighting all its neighbors with standard normal density kernel K (z/h), where z was defined as the distance between the point and center of each image patch, and h, the bandwidth, was defined as 2 times the estimated cell diameter. A region with probability larger than 0.5 was defined as a tumor cell region. The same approach was used to define stromal cell region and lymphocyte cell region. Next, ConvPath calculated 6 features for each region (**Supplemental Table 3**), which were the perimeter divided by the square root of region area and size divided by region area for the 3 kinds of cell regions separately.

**Statistical analysis**

R (version 3.2.4)[29] and R packages survival (version 2.38-3), glmnet (version 2.0-5), and clinfun (version 1.0.13) were used for statistical analysis. Survival time was defined as period from diagnosis to death or last contact for the NLST and TCGA datasets, and from diagnosis to recurrence or last contact in the CHCAMS dataset. The prognostic model was trained on the NLST patients using a Cox regression model with elastic penalty, to predict a risk score for each sampling region. The final risk score of each patient was determined by averaging risk scores across 10 sampling regions of this patient. The performance of this prognostic model was evaluated on the TCGA and CHCAMS datasets by dichotomizing the patients by the median predicted risk score of each dataset. In the validation study, the maximum follow-up time was set to six years, since the patient survival after six year may not directly relate to cancer specific events. Kaplan-Meier (K-M) plots and log rank tests were used to compare survival outcomes. In addition, a multivariate Cox proportional hazard model was used to test whether the prognostic risk scores were statistically significant after adjusting for other clinical variables, including age, gender, tobacco history, and stage. A Jonckheere-Terpstra (J-T) k-sample test[30] was used to test whether higher risk scores were correlated with theoretically more severe adenocarcinoma subtypes. The results were considered significant if the two-sided (except for J-T test, which is one-sided) test p value ≤ 0.05.

**Data availability**

Pathology images and clinical data in the NLST and TCGA datasets that support the findings of this study are available online in the NLST (https://biometry.nci.nih.gov/cdas/nlst/) and The Cancer Genome Atlas Lung Adenocarcinoma (TCGA-LUAD, https://wiki.cancerimagingarchive.net/display/Public/TCGA-LUAD). Data in the SPORE and CHCAMS datasets that support the findings of this study are available from the UT Lung SPORE Tissue bank and the National Cancer Center/Cancer Hospital, Chinese Academy of Medical Sciences and Peking Union Medical College (CHCAMS), China, separately, but restrictions apply to the availability of these data.

**Code availability**

The codes are publically accessible via https://qbrc.swmed.edu/projects/cnn/.

**RESULTS**

**ConvPath classifies lung adenocarcinoma cell types with high accuracy**

11,988 tumor, stromal, and lymphocyte image patches centering at cell nuclei centroids were extracted from region of interests (ROIs) in the TCGA and NLST datasets (**Figure 2, Supplemental Table 3**) and used to train the CNN model (**Figure 3a**). ROIs are regions of the slides which contain the majority of malignant tissues (**Figure 2a**). Example image patches are shown in **Supplemental Figure 1**. The overall classification accuracies of the CNN model on training images were 99.3% for lymphocytes, 87.9% for stroma cells, and 91.6% for tumor cells, respectively (**Figure 3b**). The independent cross-study classification rates in the SPORE dataset were 97.8% for lymphocytes, 86.5% for stroma cells, and 85.9% for tumor cells (**Figure 3c**).

**Tumor micro-environment features from predicted sampling regions correlate with overall survival**

ConvPath was then used to generate cell type predictions for 10 random sampling regions within the ROI on each slide. Based on nuclei centroid locations together with accurate cell type predictions (**Figure 4a, Supplemental Figure 2**), we investigated whether spatial distributions of tumor cells, stromal cells, and lymphocytes correlated with the survival outcome of lung ADC patients. In each predicted sampling region, tumor, stromal, and lymphocyte cell regions were detected using a kernel smoothing algorithm (**Figure 4b, Method section**). For regions of each cell type, simple parameters such as perimeter and size were measured. To ensure comparability across image slides captured at different magnitudes, the parameters were normalized by area of sampling region. In univariate Cox analysis, 4 of the 6 extracted features significantly correlated with survival outcome in the NLST dataset (**Supplemental Table 2**). Interestingly, both perimeter and area of stroma region were good prognostic factors, suggesting a protective effect of stroma cells in lung ADC patients (**Supplemental Figure 3**).

**Development and validation of an image feature-based prognostic model**

Utilizing the region features of each cell type extracted from the pathology images in the NLST dataset, we developed a prognostic model to predict patient survival outcome (coefficients of this model are shown in **Supplemental Table 2**). The model was then independently validated in the TCGA and CHCAMS datasets. The TCGA and CHCAMS patients were dichotomized according to the median predicted risk scores in each dataset. In both datasets, the patients in the predicted high-risk group had significantly worse survival outcome than those in the predicted low-risk group (**Figure 5 a&b**, log rank test, $p = 0.0047$ for the TCGA dataset, $p = 0.030$ for the

CHCAMS dataset). To evaluate whether the image features extracted by ConvPath were independent of clinical variables, multivariate Cox proportion hazard models were used to adjust the predicted risk scores with available clinical variables, including gender, age, stage and smoking status (**Table 1**). After adjustment, the still significant hazard ratios between high- and low-risk groups (p = 0.0021 for the TCGA dataset, p = 0.016 for the CHCAMS dataset) indicated that risk group as defined by ConvPath-extracted image features was an independent prognostic factor, in addition to other clinical variables.

**Predicted risk scores correlate with severity of ADC subtypes**

The 2015 WHO classification of lung cancer further divides invasive lung ADC into several subtypes, including acinar, lepidic, micropapillary, papillary, solid, and mucinous ADC[2]. The correlation of the predicted risk scores with predominant histology subtypes identified by our pathologist for the CHCAMS dataset, according to the 2015 WHO classification guidelines, was tested (**Figure 5c**). Higher risk scores correlated with more aggressive ADC subtypes, such as solid predominant ADC and invasive mucinous ADC (p = 0.0039).

**The ConvPath software and web server**

To facilitate practical application of this pathological image analysis pipeline by pathologists and bioinformaticians, the image segmentation, deep learning, and feature extraction algorithms were incorporated into the ConvPath software. The ConvPath software is publicly accessible from the web server created for this study, which is at https://qbrc.swmed.edu/projects/cnn/ (**Supplemental Figure 6**).

**DISCUSSION**

In this study, an image analysis and cell classification pipeline was developed. It can perform nuclei segmentation, CCN based cell type prediction, and feature extraction (**Figure 1**). This pipeline successfully visualizes the spatial distributions of tumor, stromal, and lymphocyte cells in ROI of lung ADC pathology images. It can potentially serve as a prognostic method independent of other clinical variables. The patient prognostic model based on extracted image features was trained in the NLST dataset and independently validated in the TCGA and CHCAMS datasets, which indicates the generalizability of this analysis pipeline to other lung ADC patients.

The accurate classification of cell types in pathology images was validated in an independent data cohort. While the qualities of H&E staining vary across different cohorts and there are inherent inter-patient differences, ConvPath still has 90.1% overall accuracy in the SPORE dataset (**Figure 3c**). The robustness of ConvPath benefits from the level set-based segmentation algorithm in the nuclei segmentation step. This segmentation algorithm is invariant to the location of initial contour and can handle high variability across different H&E pathology images. Moreover, nuclei centroid extraction based on distance transform can separate most of the connected nuclei that are not properly processed by the commonly used CellProfiler software[19, 31]. The robustness of prediction also benefits from the powerful CNN, which is designed to emulate the behavior of the visual cortex, and poses properties of deep structure, local connectivity, and shared weights.

The relationships between the extracted tumor micro-environment-related image features and patient prognosis were evaluated in this study (**Supplemental Table 2**). In univariate analysis, higher stromal cell abundance correlated with better prognosis (**Supplemental Figure 4**), which is consistent with a recent report on lung ADC patients[10]. However, controversial roles of stroma cells in tumor progression have been reported, including stimulation of tumor proliferation through growth signals and limitation of tumor cells metastatic spreading[8, 9, 32]. Combinatory analysis of cell spatial distribution detected in this study and the functionality of stroma cells, which could not be evaluated through H&E staining, will help answer whether these controversial roles arise from the different activation status of crosstalk between tumor and stroma. In contrast, higher lymphocyte abundance, reflected by region size rather than perimeter, correlated with worse prognosis (**Supplemental Table 2, Supplemental Figure 5**). However, although the presence of both tumor- and stroma-infiltrating lymphocytes has been reported to correlate with tumor cell apoptosis and better patient survival in non-small cell lung cancer[11, 14, 33], the tumor-suppressive or tumor-promoting properties of lymphocytes depend on the tumor microenvironment[34]. Thus, quantifying distribution and interaction with tumor or stroma cells of lymphocytes can potentially provide a way to evaluate immune response status and serve as a biomarker for immunotherapy response.

The goal of this study is to develop software tools to automatically identify cells and classify cell types in tumor pathology image. Since there are more than 10,000 cells in each sampling region (**Supplemental Figure 2**), it is extremely labor-intensive and error-prone to manually localize and label each cell nuclei. Thus, automatic visualization of distributions of different cell types will facilitate the diagnostic procedure. More importantly, extracting image features directly from the labeled sampling region allows for accurate quantification of tumor, stroma, and lymphocyte regions, which can avoid subjective assessment by human pathologists. In addition, this study provides an automatic and quantitative tool to dissect intra-tumor heterogeneity on the cell-type level, which has been reported to inform metastasis[35], immunotherapy responsiveness[36], and angiogenesis inhibitor responsiveness[37]. The analysis pipeline developed in this study could convert the pathology image into a "spatial map" of tumor cells, stromal cells and lymphocytes. This will greatly facilitate and empower comprehensive spatial analysis of cell distributions and interactions, as well as their roles in tumor progression and metastasis.

There are several limitations of the ConvPath pathology image analysis pipeline. First, the sampling region selection and subsequent steps rely on ROI labeling, which is currently done by pathologists. We are working on fully automated ROI selection, which will further decrease the bias caused by subjectivity. Second, only three major cell types are considered in the ConvPath CNN algorithm; thus, this CNN model is sensitive to out-of-focus cell types such as macrophages and epithelial cells. Also, different subtypes of lymphocytes, such as CD4+ and CD8+ T cells, are not distinguishable using our algorithm[11, 38]. More comprehensive labeling and immunohistochemical staining will help solve this problem. Third, more comprehensive analysis of spatial distribution of cells is not included in this research[37, 39]. Analyzing the spatial patterns, such as cell clustering and inter-cell interactions, will help us understand the mechanism of tumor progression and immune response to tumor cells.

**Author Contributions:**

G.X. and T.W. supervised the project. S.W., T.W., L.Y. and G.X. conceived the method. S.W., T.W., L.Y. and G.X. designed and performed the analyses and interpreted the results. L.Y., Y.Y, J.F. I.W., Y.M. and Y.X. collected and provided the data. S.W., T.W., L.Y., F.Y, X.L., Y.Y. and A, G. curated the data. C.L., S.W., S.L., and B.Y developed the web application with advice from G.X., T.W. and Y.X.

L.Y., A.G., J.F., and I.W. provided critical input. S.W. and T.W. drafted the article. All co-authors have read and edited the manuscript.

**Competing financial interests:**

The authors declare that they have no competing interests.

**Materials & Correspondence:**

Correspondence and material requests should be addressed to G.X.

**Source of Funding:**

This work was supported by the National Institutes of Health [1R01GM115473, 5R01CA152301, 5P30CA142543 and 1R01CA172211); and the Cancer Prevention and Research Institute of Texas [RP120732].

**Table 1.** Multivariate analysis of the predicted risk scores in the CHCAMS and TCGA datasets adjusted by clinical variables.

| TCGA dataset (n=346) | HR | 95% CI | p value |
|---|---|---|---|
| High risk vs. low risk | 2.19 | (1.33-3.60) | 0.0021 |
| Age (per year) | 1.03 | (1.01-1.06) | 0.014 |
| Male vs. female | 0.69 | (1.45-1.16) | 0.16 |
| Smoker vs. non-smoker | 0.88 | (0.53-1.47) | 0.62 |
| Stage | | | |
|   Stage I | ref | | - |
|   Stage II | 2.69 | (1.45-5.00) | 0.0017 |
|   Stage III | 5.04 | (2.69-9.43) | <0.001 |
|   Stage IV | 6.06 | (2.49-14.73) | <0.001 |
| **CHCAMS dataset (n=88)** | HR | 95% CI | p value |
| High risk vs. low risk | 2.21 | 1.16-4.21 | 0.016 |
| Age (per year) | 1.02 | 0.99-1.06 | 0.202 |
| Male vs. female | 1.85 | 0.69-4.91 | 0.22 |
| Smoker vs. non-smoker | 0.76 | 0.28-2.04 | 0.585 |

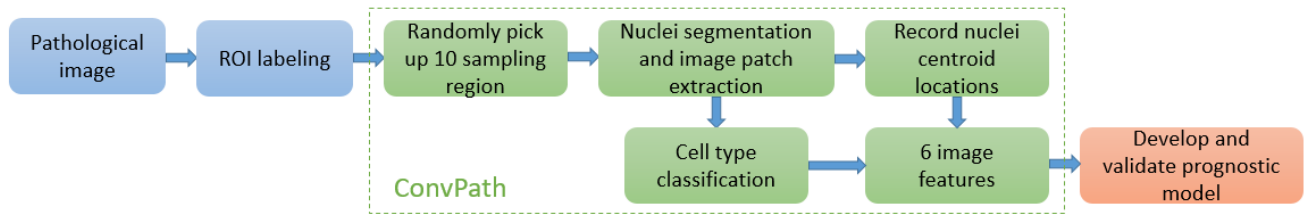

CHCAMS, National Cancer Center/Cancer Hospital of Chinese Academy of Medical Sciences, China; CI, confidence interval; HR, hazard ratio; TCGA, The Cancer Genome Atlas. **Figure 1.** Flow chart of ConvPath-aided pathological image analysis.

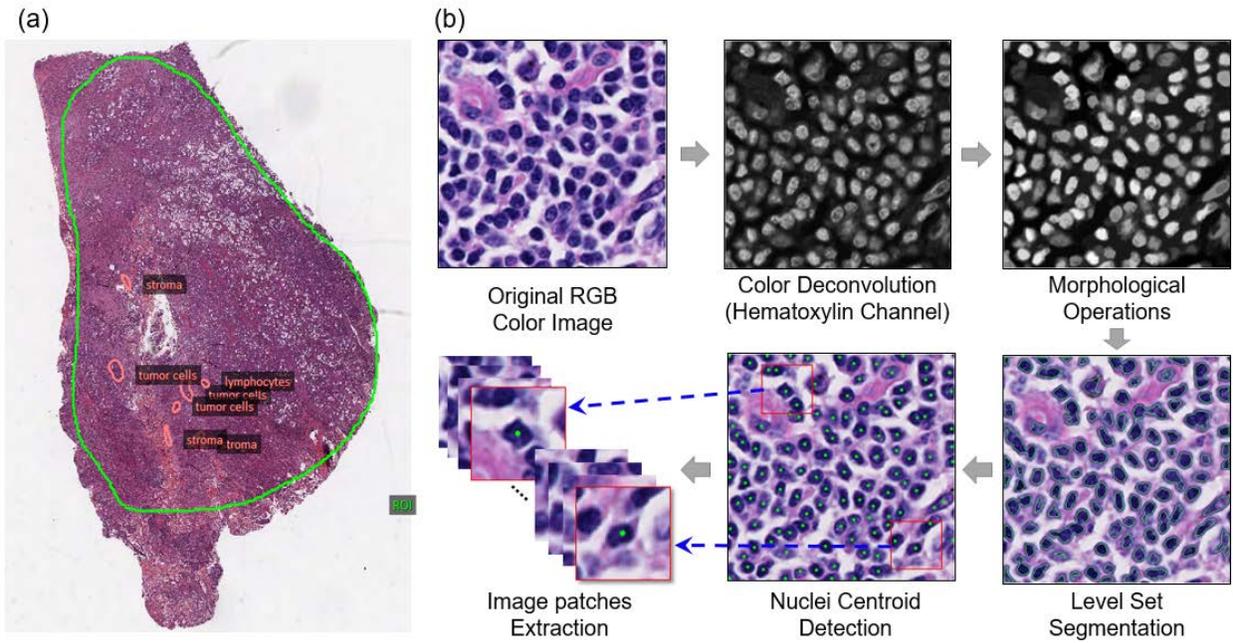

**Figure 2.** Image preprocessing step of the ConvPath software. **(a)** Selection of regions of interest (ROIs) in whole pathological imaging slides. **(b)** Image segmentation pipeline to extract cell-centered image patches from selected ROIs.

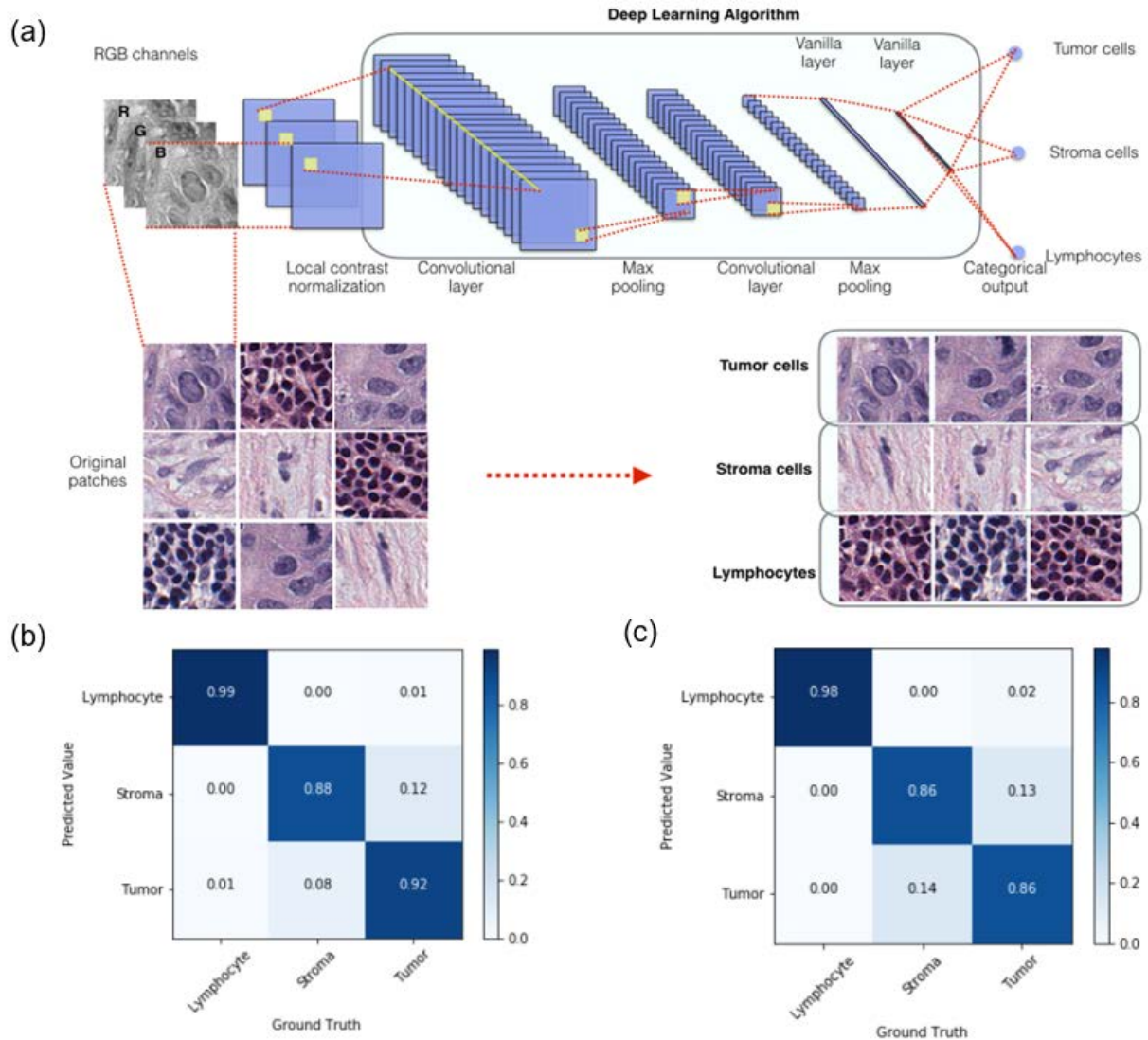

**Figure 3.** Cell type recognition step of the ConvPath software. **(a)** Schema and structure of the convolutional neural network (CNN) to recognize the types of cells in the centers of image patches. **(b)** Confusion matrix of internal testing results of CNN on the NLST and TCGA training image slides. Prediction accuracies are calculated based on 3996 image patches for each cell type. **(c)** Confusion matrix of independent testing results of CNN on image patches of the SPORE dataset. Prediction accuracies are calculated based on 8245 lymphocyte, 2211 stroma, and 6836 tumor patches.

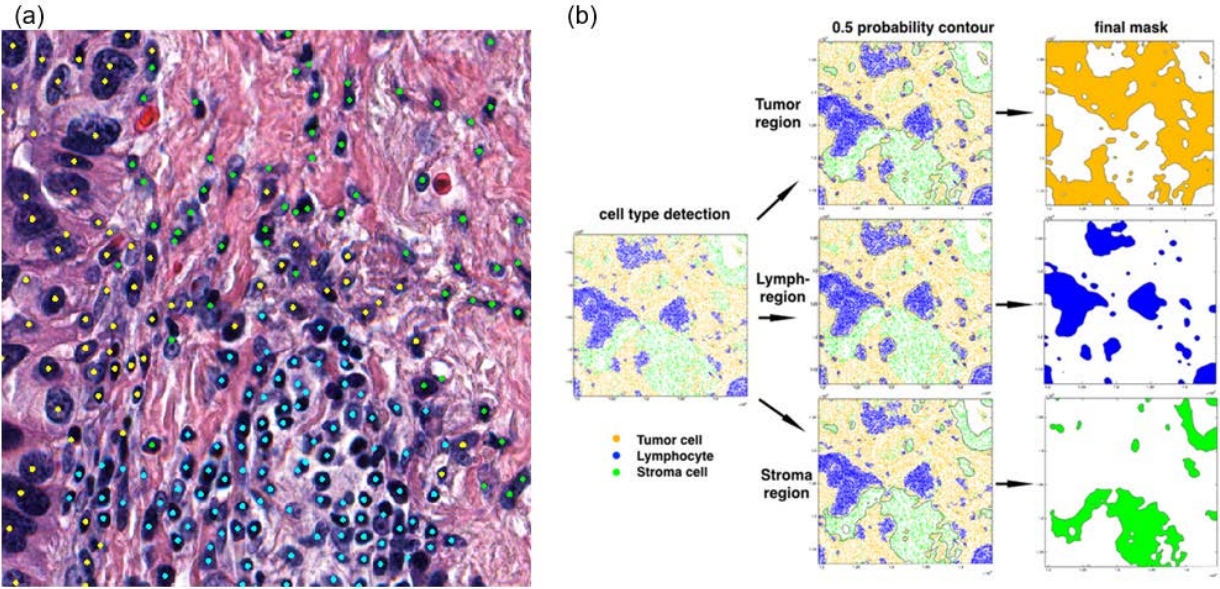

**Figure 4.** Feature extraction step of the ConvPath software. **(a)** A zoomed-in part of a sampling region (**Supplemental Figure 3**) in which cell nuclei centroids are labeled with predicted cell types. Green, stroma; cyan, lymphocyte; yellow, tumor. **(b)** Cell type region detection using a kernel smoothing algorithm for the sampling region shown in **Supplemental Figure 3**. Area and perimeters are evaluated for regions of tumor, stroma, and lymphocyte.

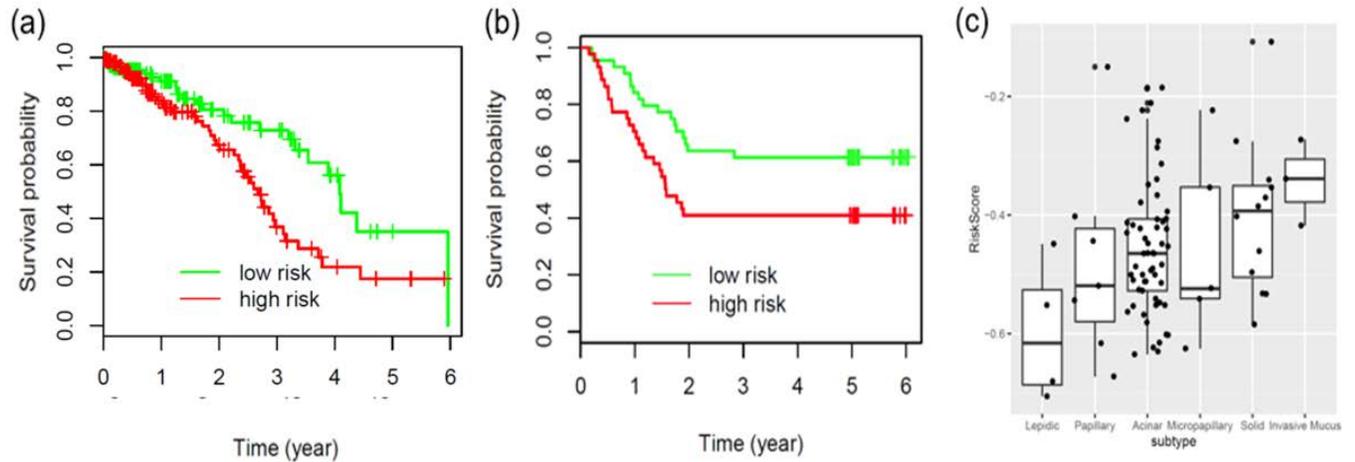

**Figure 5.** Application of the prognostic model to independent datasets. **(a, b)** Validation of the prognostic model in the TCGA overall survival data (a, log rank test, p = 0.0047) and the CHCAMS recurrence data (b, log rank test, p = 0.030). **(c)** Distribution of predicted risk scores in the 5 histological subtypes of lung adenocarcinoma for the CHCAMS dataset patients. Jonckheere-Terpstra k-sample test, p = 0.0039.

# SUPPLEMENTAL MATERIAL

**Supplemental Table1.** Patient population characteristics for TCGA, NLST, CHCAMS and SPORE datasets.

| Cohort | | TCGA | NLST | Beijing | SPORE |
|---|---|---|---|---|---|
| **Number of patients** | | 523 | 201 | 102 | 112 |
| **Number of slides (Tumor)** | | 1337 | 345 | 102 | 130 |
| **Age at diagnosis (years, median [LQ-HQ])** | | 66 [59-73] | 64 [60-68] | 59 [54-66] | 65 [58-73] |
| **Follow-up (years, median [LQ-HQ])** | | 0.6 [0.1-2.0] | 6.6 [5.4-6.9] | 5.0 [4.1-5.8] | 3.3 [1.7-5.3] |
| **Vital status (%)** | Alive | 394 (75.3) | 136 (67.7) | 74 (72.5) | 75 (67.0) |
| | Deceased | 126 (24.1) | 64 (31.8) | 28 (27.5) | 37 (33.0) |
| | NA | 3 (0.6) | 1 (0.5) | 0 (0.0) | 0 (0.0) |
| **Gender (%)** | M | 242 (46.3) | 112 (55.7) | 46 (45.1) | 56 (50.0) |
| | F | 278 (53.2) | 89 (44.3) | 56 (54.9) | 56 (50.0) |
| | NA | 3 (0.6) | 0 (0.0) | 0 (0.0) | 0 (0.0) |
| **Cancer stage (%)** | I | 135 (67.7) | 135 (67.2) | 102 (100.0) | 70 (62.5) |
| | II | 20 (9.9) | 20 (10.0) | 0 (0.0) | 17 (15.2) |
| | III | 33 (16.4) | 33 (16.4) | 0 (0.0) | 24 (21.4) |
| | IV | 13 (6.5) | 13 (6.5) | 0 (0.0) | 1 (0.9) |
| | NA | 4 (0.8) | 0 (0.0) | 0 (0.0) | 0 (0.0) |
| **Smoking status (%)** | Smoker | 431 (82.4) | 110 (54.7) | 43 (42.2) | 98 (87.5) |
| | Non-smoker | 75 (14.3) | 91 (45.3) | 59 (57.8) | 13 (11.6) |
| | NA | 17 (3.3) | 0 (0.0) | 0 (0.0) | 1 (0.9) |

CHCAMS, National Cancer Center/Cancer Hospital of Chinese Academy of Medical Sciences, China; HQ, higher quantile at 75%; LQ, lower quantile at 25%; NLST, the National Lung Screening Trial; TCGA, The Cancer Genome Atlas.

**Supplemental Table 2.** Breakdown of the numbers of image patches from each training dataset for the deep learning algorithm in ConvPath.

| Data source | Cell type | # image patch |
| --- | --- | --- |
| NLST | lymphocyte | 2096 |
| NLST | stroma | 2550 |
| NLST | tumor | 1298 |
| TCGA | lymphocyte | 1900 |
| TCGA | stroma | 1446 |
| TCGA | tumor | 2698 |

NLST, the National Lung Screening Trial; TCGA, The Cancer Genome Atlas.

**Supplemental Table 3.** Extracted cell type-level image features with their data distribution, explanation, and univariate analysis results in the NLST dataset.

| Features | Univariate in the NLST dataset | | Glmnet model | Data range, median (min - max) | | |
|---|---|---|---|---|---|---|
| | HR | p value | Coef. ($\lambda = 0.02$*) | NLST | TCGA | CHCAMS |
| Perimeter of lymphocyte cell region/square root of image size† | 1.006 | 0.30 | -0.0019 | 4.92 (0 - 23.04) | 0.2 (0 - 18.5) | 0.14 (0 - 13.15) |
| Perimeter of stromal cell region/square root of image size | 0.97 | < 0.001 | -0.015 | 6.47 (0 - 23.51) | 9.95 (0 - 26.62) | 13.38 (0 - 25.24) |
| Perimeter of tumor cell region/square root of image size | 0.98 | 0.0016 | -0.014 | 13.4 (0 - 26.6) | 11.99 (0 - 26.04) | 14 (0 - 26.61) |
| Size of lymphocyte cell region/image size | 1.63 | 0.0012 | 0.26 | 0.08 (0 - 1) | 0 (0 - 0.72) | 0 (0 - 1) |
| Size of stromal cell region/image size | 0.53 | < 0.001 | -0.16 | 0.1 (0 - 0.94) | 0.19 (0 - 1) | 0.3 (0 - 0.98) |
| Size of tumor cell region/image size | 0.96 | 0.71 | -0.060 | 0.61 (0 - 1) | 0.66 (0 - 1) | 0.55 (0 - 1) |

CHCAMS, National Cancer Center/Cancer Hospital of Chinese Academy of Medical Sciences, China; Coef., coefficient; HR, hazard ratio; NLST, the National Lung Screening Trial; TCGA, The Cancer Genome Atlas.

* $\lambda$ is the penalty coefficient in the glmnet model. Its value has been optimized by 10-fold cross-validation.

† Image here refers to the square sampling region.

## Lymphocytes

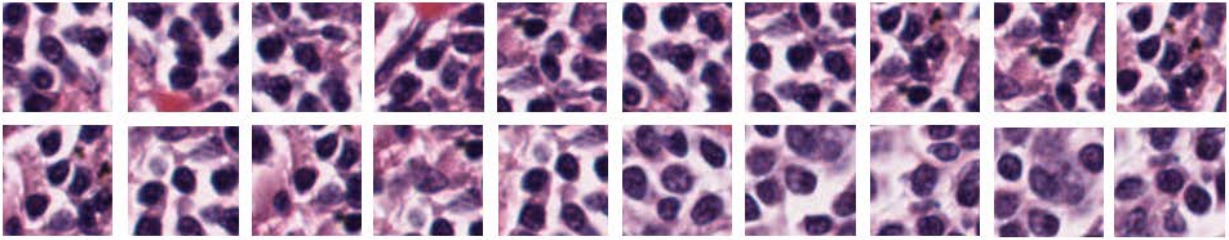

## Stroma cells

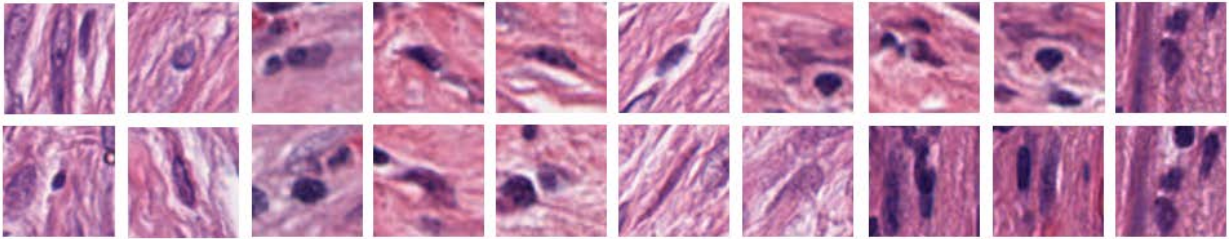

## Tumor cells

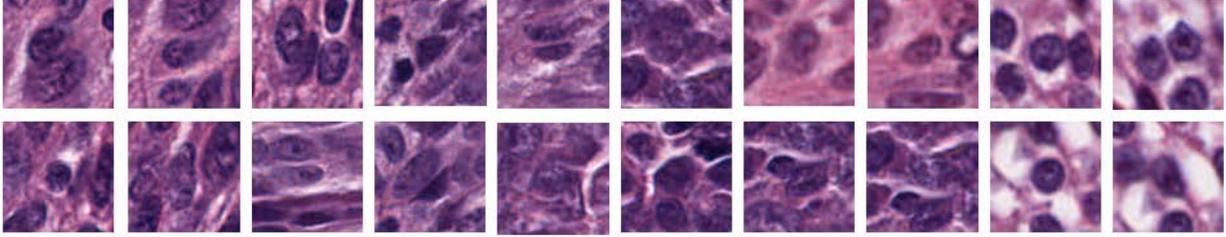

**Supplemental Figure 1.** Example 80×80 pixels image patches centering at cell nuclei centroids.

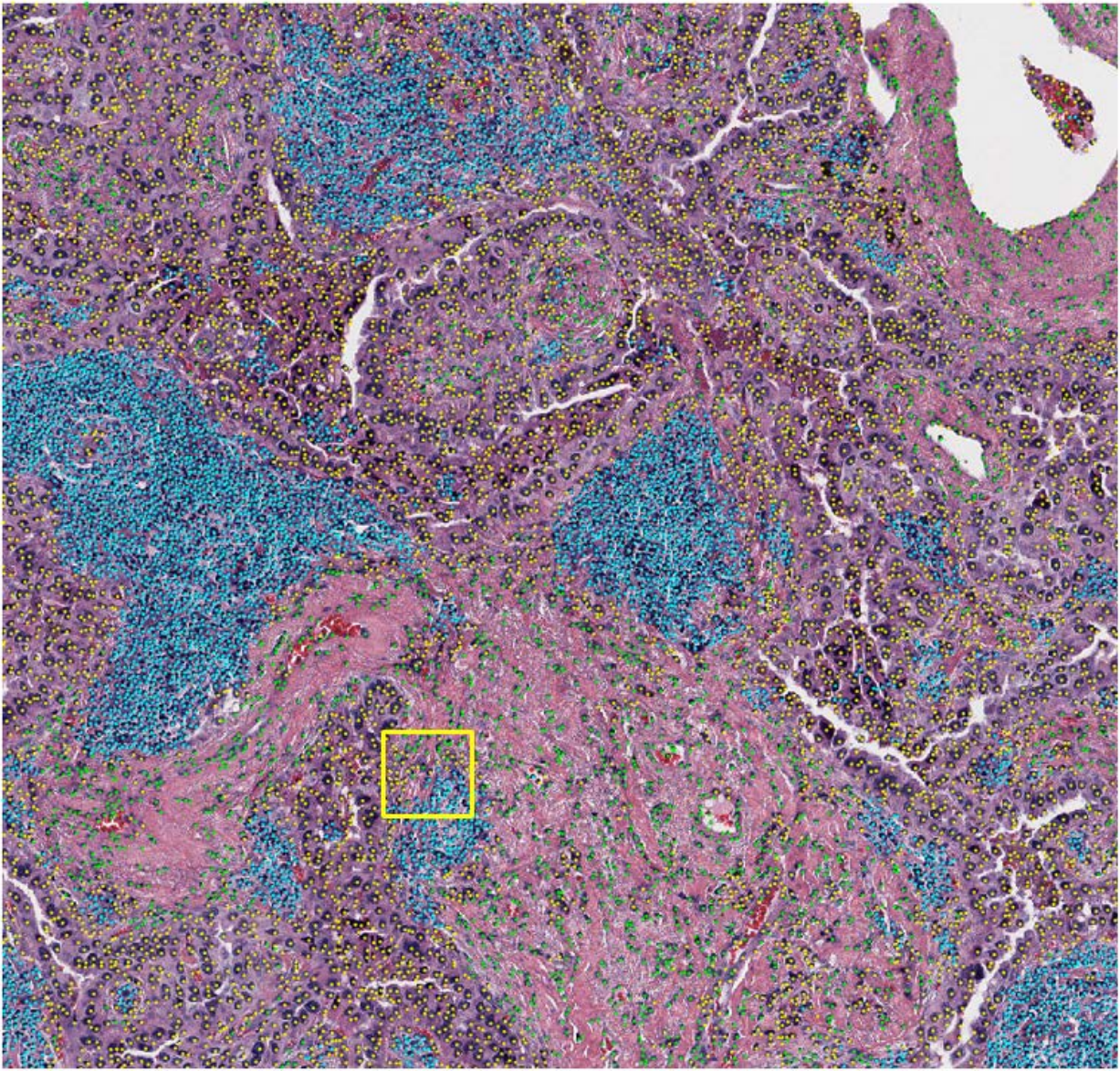

**Supplemental Figure 2.** A sampling region in H&E-staining slide on which all cells are color-labeled by CNN predictions. Green, stroma; cyan, lymphocyte; yellow, tumor. The red rectangle is enlarged and shown in **Figure 4a**.

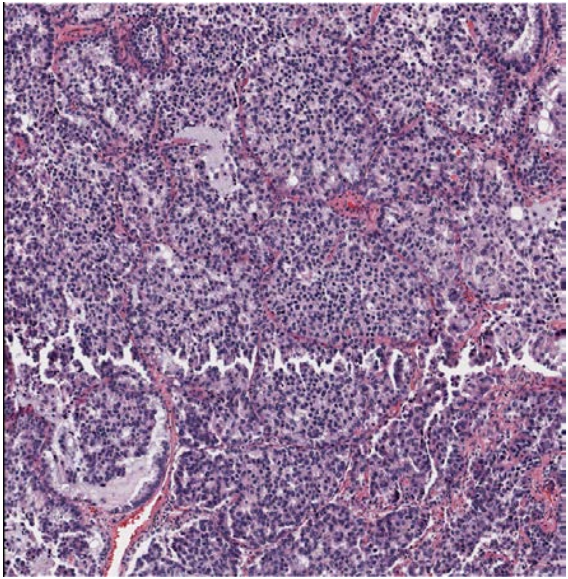
(a)

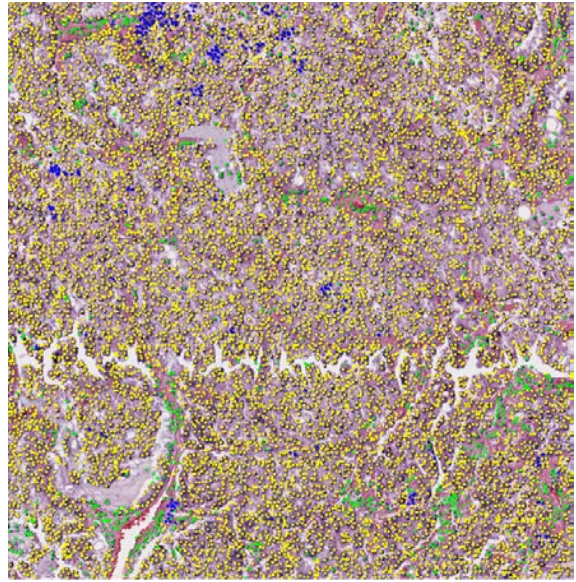
(b)

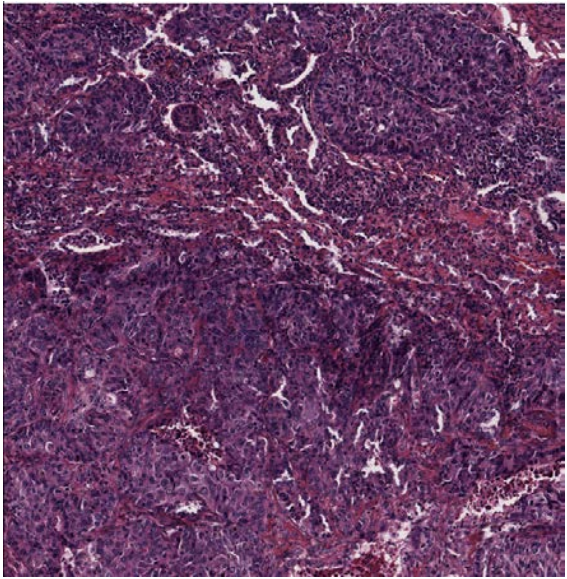
(c)

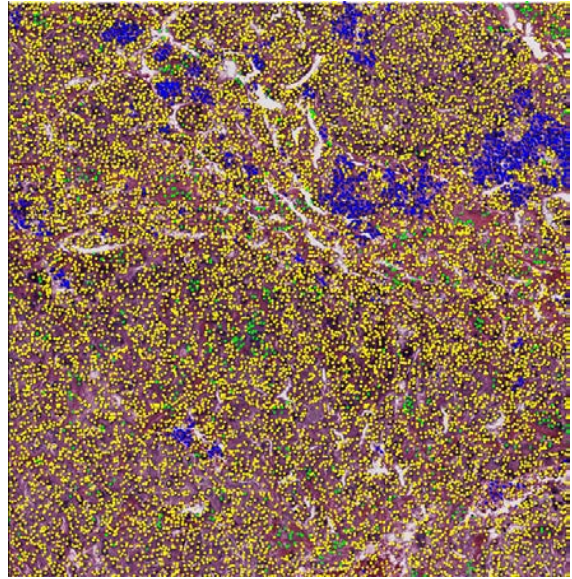
(d)

**Supplemental Figure 3.** Examples of original sampling regions (a&c, left panels) and the regions labeled by ConvPath (b&d, right panels). Exemplar images in which there are few stroma cells or lymphocytes. Yellow, tumor cells; green, stroma cells; blue, lymphocytes.

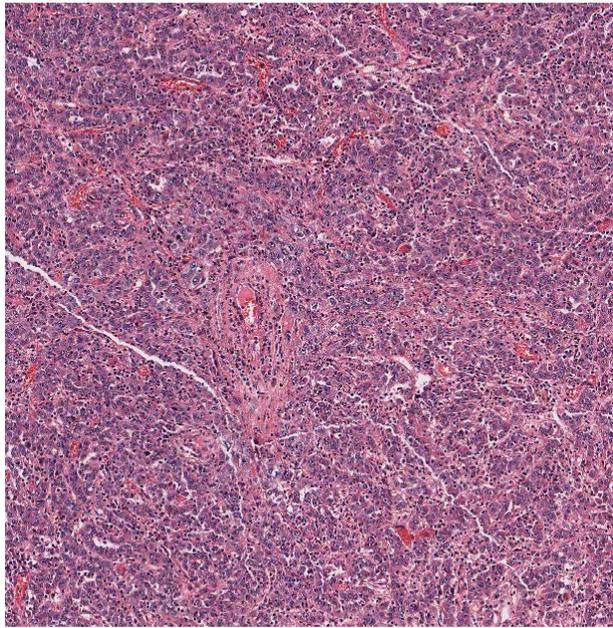 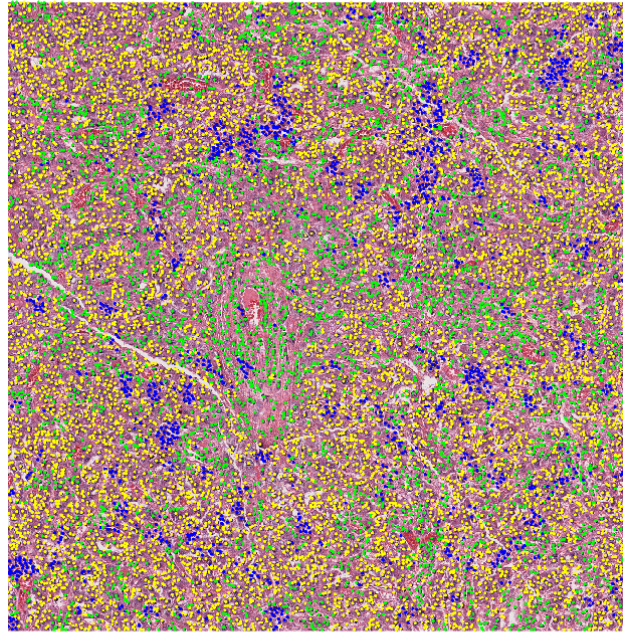
(a) (b)

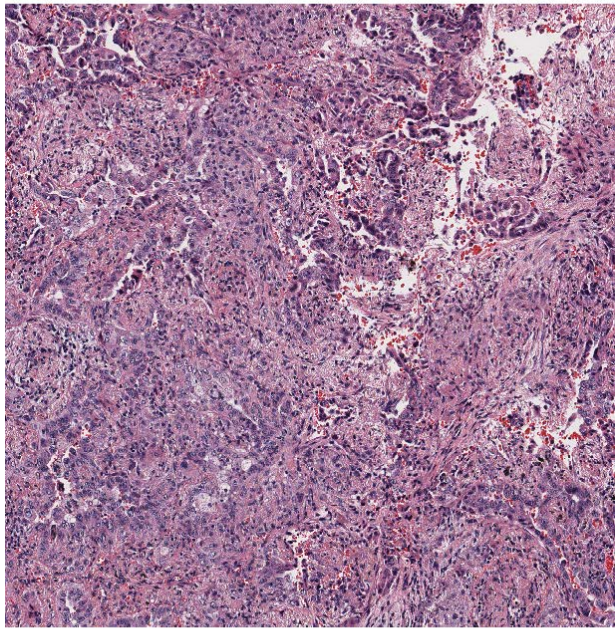 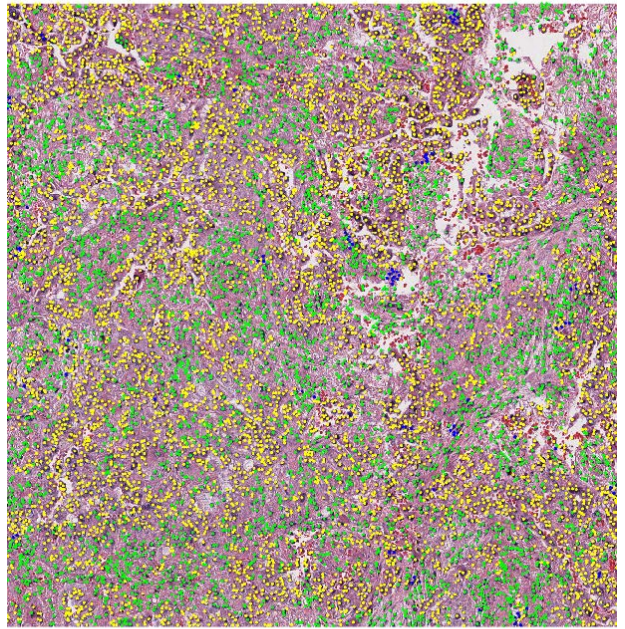
(c) (d)

**Supplemental Figure 4.** Examples of original sampling regions (a&c, left panels) and the regions labeled by ConvPath (b&d, right panels). Exemplar images in which tumor cells are surrounded with stroma cells. Yellow, tumor cells; green, stroma cells; blue, lymphocytes.

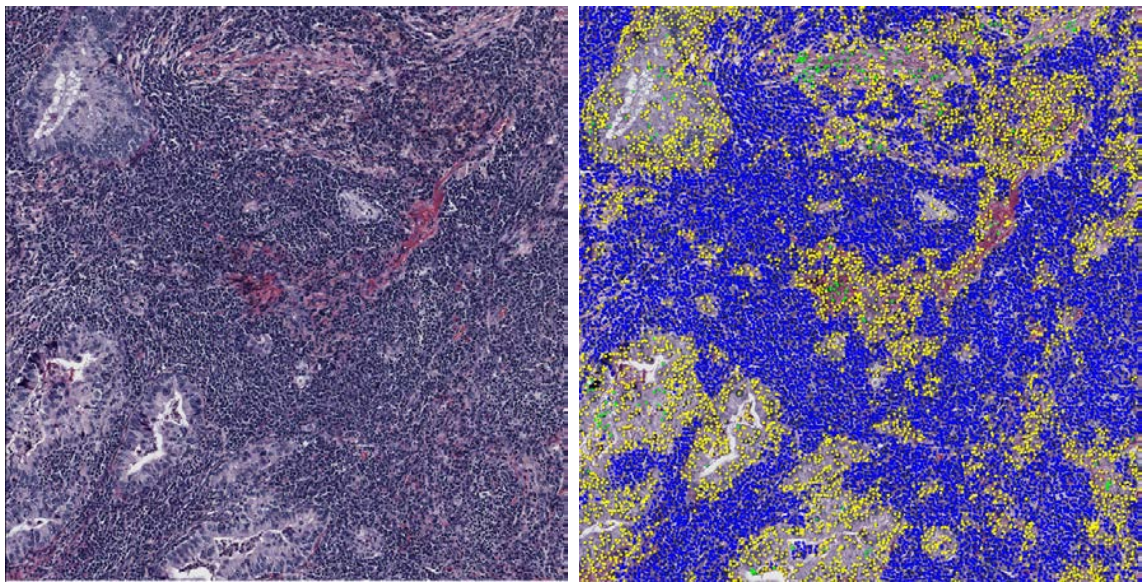

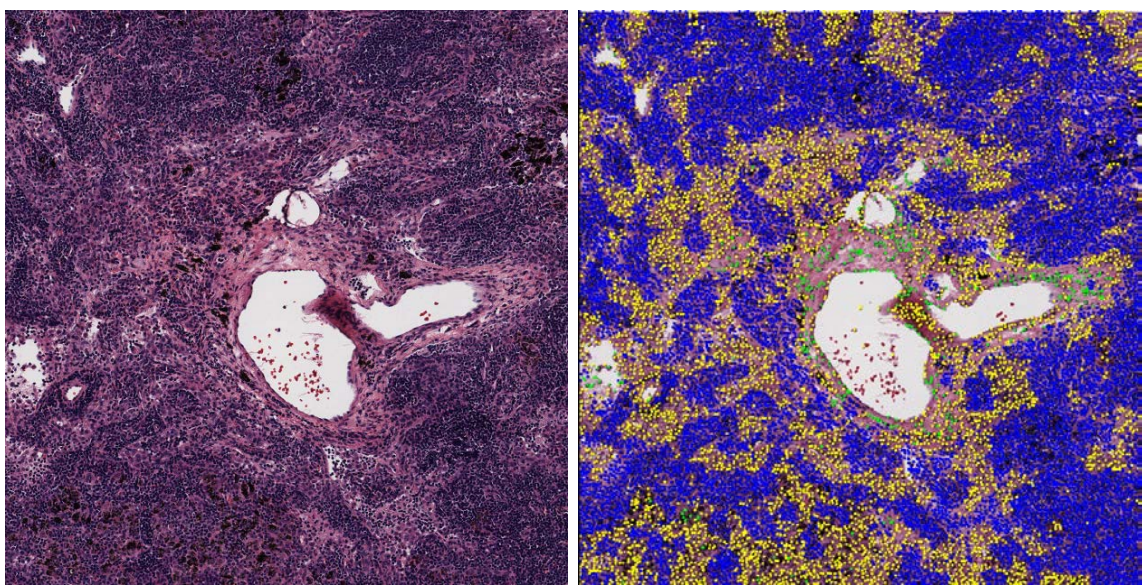

**Supplemental Figure 5.** Examples of original sampling regions (a&c, left panels) and the regions labeled by ConvPath (b&d, right panels). Exemplar images in which there are lymphocyte infiltration. Yellow, tumor cells; green, stroma cells; blue, lymphocytes.

**Supplemental Figure 6.** Screenshot of the webserver from which potential users can download source codes, sample test data, and user manual explaining usage of ConvPath. (https://qbrc.swmed.edu/projects/cnn/).